\documentclass[final]{cvpr}

\usepackage{times}
\usepackage{epsfig}
\usepackage{graphicx}
\usepackage{amsmath}
\usepackage{amssymb}
\usepackage{multirow}
\usepackage{color}
\usepackage{booktabs}
\usepackage{bbding}
\usepackage{makecell}

\usepackage[pagebackref=true,breaklinks=true,colorlinks,bookmarks=false]{hyperref}

\begin{document}

\title{PP-LCNet: A Lightweight CPU Convolutional Neural Network}  % **** Enter the paper title here

\author{Cheng Cui, Tingquan Gao, Shengyu Wei, Yuning Du, \\
Ruoyu Guo, Shuilong Dong, Bin Lu, Ying Zhou, Xueying Lv,  \\
Qiwen Liu, Xiaoguang Hu, Dianhai Yu, Yanjun Ma \\
Baidu Inc.\\
\tt\small \{cuicheng01, gaotingquan, weishengyu, duyuning\} @baidu.com
}

\maketitle
\thispagestyle{empty}

\begin{abstract}

We propose a lightweight CPU network based on the MKLDNN acceleration strategy, named PP-LCNet, which improves the performance of lightweight models on multiple tasks. This paper lists technologies which can improve network accuracy while the latency is almost constant. With these improvements, the accuracy of PP-LCNet can greatly surpass the previous network structure with the same inference time for classification. As shown in Figure \ref{Figure1}, it outperforms the most state-of-the-art models. And for downstream tasks of computer vision, it also performs very well, such as object detection, semantic segmentation, etc. All our experiments are implemented based on PaddlePaddle$\footnote{https://github.com/PaddlePaddle}$. Code and pretrained models are available at PaddleClas$\footnote{https://github.com/PaddlePaddle/PaddleClas}$.

\end{abstract}

%%%%%%%%% BODY TEXT - ENTER YOUR RESPONSE BELOW
\section{Introduction}

In the past few years, Convolutional Neural Networks (CNNs) represent the workhorses of the most current computer vision applications, including image classification\cite{alexnet, urnet}, object detection\cite{fasterrcnn}, attention prediction\cite{attentiongrad}, target tracking\cite{targettracking}, action recognition\cite{action}, semantic segmentation\cite{deeplab,deeplabv3+}, salient object detection\cite{SOD} and edge detection\cite{edgedetect}.

As the model feature extraction capability increases and the number of model parameters and FLOPs get larger, it becomes difficult to achieve fast inference speed on mobile devices based ARM architecture or CPU devices based x86 architecture. In this case, many excellent mobile networks have been proposed, but due to the limitations of the MKLDNN, the speed of these networks is not ideal on the Intel CPU with MKLDNN enabled. In this paper, we rethink the lightweight models elements for network designed on Intel-CPU. In particular, we consider the following three fundamental questions. (i) How to promote the network to learn stronger feature presentations without increasing latency.  (ii) What are the elements to improve the accuracy of lightweight models on CPU. (iii) How to effectively combine different strategies for designing lightweight models on CPU.

Our main contribution is summarizing a series of methods to improve the accuracy without increase of inference time, and how to combine these methods to get a better balance of accuracy and speed. Based on this, we come up with several general rules for designing lightweight CNNs, and provide new ideas for other researchers to build CNNs on CPU devices. Furthermore, it can provide neural architecture search researchers with new ideas when constructing the search space, so as to get better models faster. 

\begin{figure}[t]
\centering
\includegraphics[width=\linewidth]{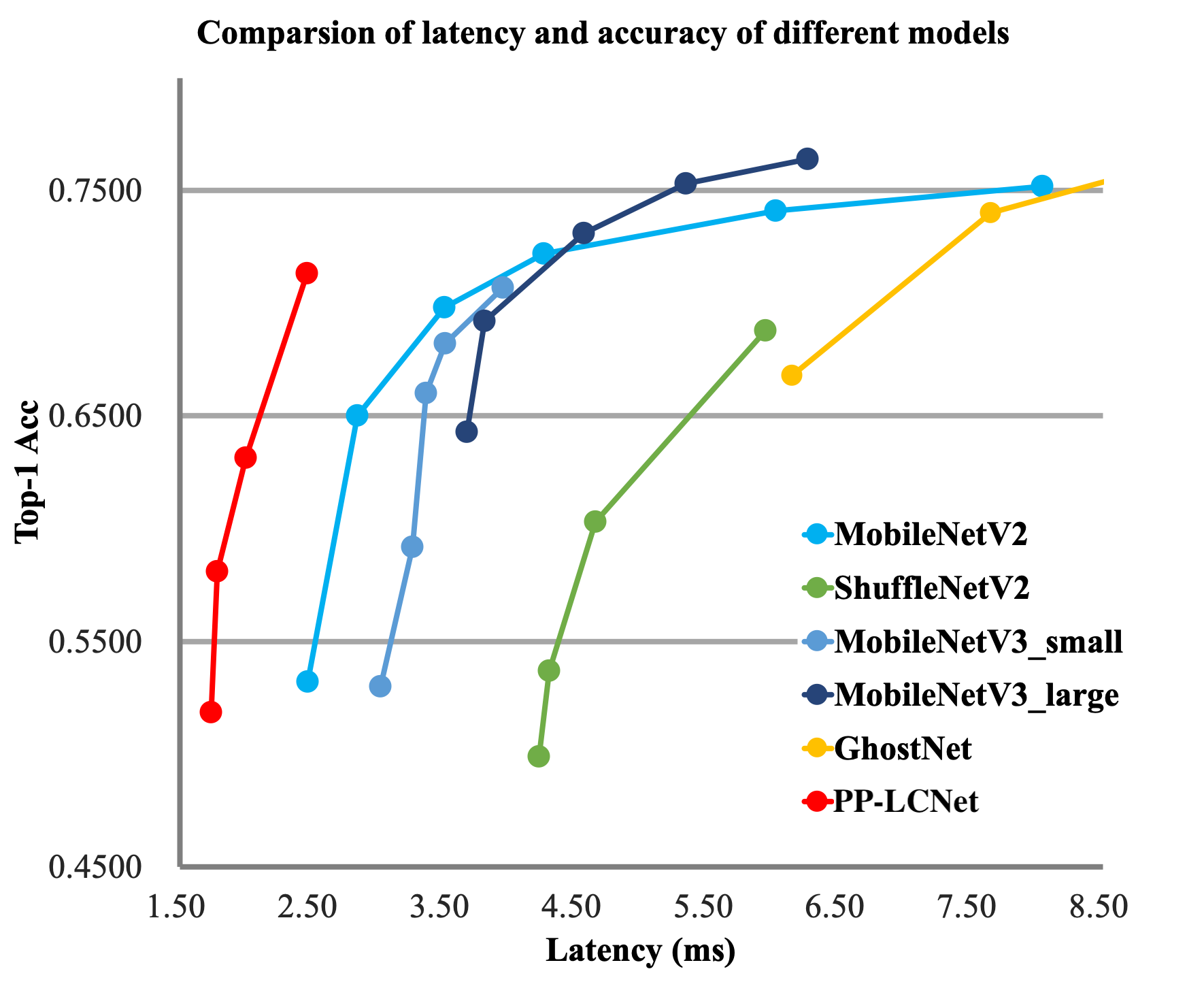} 
\caption{Comparing the accuracy-latency of different mobile series models. Latency tested on Intel$^\circledR$ Xeon$^\circledR$ Gold 6148 Processor with batch size of 1 and  MKLDNN enabled, the number of thread is 10.}
\label{Figure1}
\end{figure}

\begin{figure*}[t]
\centering
\includegraphics[width=1.0\textwidth]{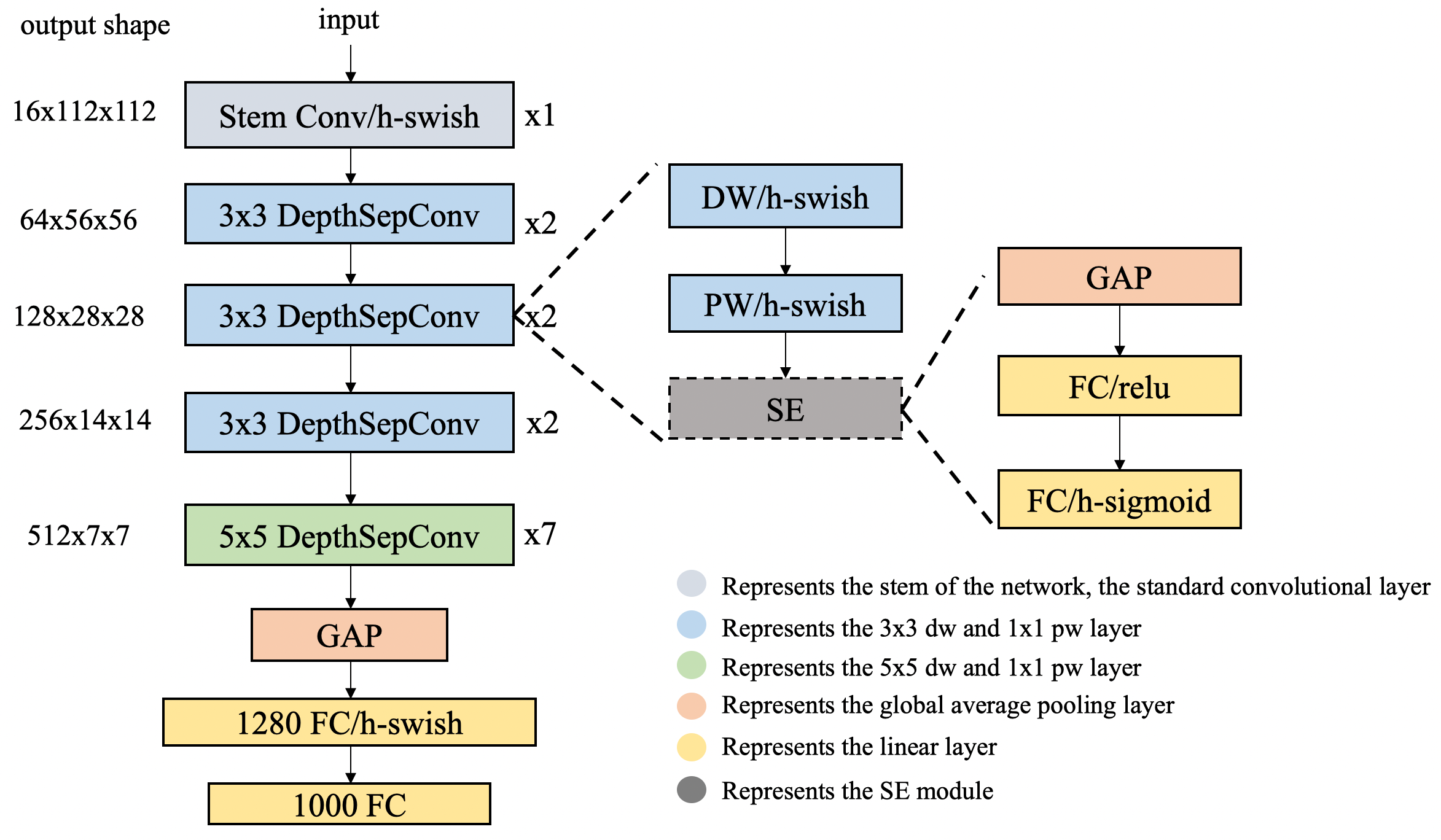} 
\caption{A detailed view of PP-LCNet. The dotted box represents optional modules.The stem part uses standard $3 \times 3$ convolution. DepthSepConv means depth-wise  separable convolutions, DW means depth-wise convolution, PW means point-wise convolution, GAP means Global Average Pooling.}
\label{Figure2}
\end{figure*}

%-------------------------------------------------------------------------

\section{Related Works}

To promote the capabilities of the model, current
works usually follow two types of methodologies. One is based on manually-designed CNN architecture, the other is based on Neural Architecture Search (NAS)\cite{zoph2016neural}.
 
 \textbf{Manually-designed Architecture.}  The VGG\cite{VGG} exhibits a simple yet effective strategy of constructing very deep networks: stacking blocks with the same dimension. GoogLeNet\cite{inceptionv1} constructs an Inception block, which includes four parallel operations: $1\times1$ convolution, $3\times3$ convolution, $5\times5$ convolution and max pooling. GoogLeNet makes the convolutional neural network light enough, then more and more lighter networks emerge. MobileNetV1\cite{mobilenetv1} replaces the standard convolution by depthwise and pointwise convolutions, which greatly reduces the amount of parameters and FLOPs of the model. The author of MobileNetV2\cite{mobilenetv2} proposed the Inverted block, which further reduces the FLOPs of the model and at the same time improves the performance of the model. ShuffleNetV1/V2\cite{shufflenet}\cite{shufflenetv2} exchanges information through channel shuffle, which reduces the unnecessary overhead of the network structure. The author of GhostNet\cite{ghostnet} proposed a novel Ghost module that can generate more feature maps with fewer parameters to improve the overall performance of the model.

\textbf{Neural Architecture Search.} With the development of GPU hardware, the main point has shifted from a manually designed architecture to an architecture that adaptively performs a systematic search for specific tasks. A majority of NAS-generated networks use the similar search space to  MobileNetV2\cite{mobilenetv2}, including EﬃcientNet\cite{efficientnet}, MobileNetV3\cite{mbv3}, FBNet\cite{fbnet}, DNANet\cite{dnanet}, OFANet\cite{ofanet} and so on. The MixNet\cite{mixnet} proposed to hybridize depth-wise convolutions of diﬀerent kernel size in one layer. NAS-generated networks relies on manually-generated block, such as BottleNeck\cite{resnet}, Inverted-block\cite{mobilenetv2} and so on. Our approach can reduce search space and improve search efficiency for neural architecture search and potentially improve the overall performance, which can be studied in future work.

\section{Approach}

\begin{table*}[!htbp]
\centering
\begin{center}
\begin{tabular}{c|c|c|c|c|c}
\toprule[1pt]
Operator & Kernel Size & Stride & Input & Output & SE \\
\midrule[1pt]
% conv1
Conv2D  & $3\times 3$     & 2     & $224^2 \times 3$  & $112^2 \times 16$ & - \\
% \hline
% blocks2
DepthSepConv     & $3\times 3$     & 1     & $112^2 \times 16$ & $112^2 \times 32$ & - \\
% blocks3_1
DepthSepConv      & $3\times 3$     & 2     & $112^2 \times 32$ & $56^2 \times 64$  & - \\
% \hline
% blocks3_2
DepthSepConv      & $3\times 3$     & 1     & $56^2 \times 64$  & $56^2 \times 64$  & - \\
% blocks4_1
DepthSepConv & $3\times 3$     & 2     & $56^2 \times 64$  & $28^2 \times 128$ & - \\

% \hline
% blocks4_2
DepthSepConv      & $3\times 3$     & 1     & $28^2 \times 128$ & $28^2 \times 128$ & - \\
% blocks5_1
DepthSepConv      & $3\times 3$     & 2     & $28^2 \times 128$ & $14^2 \times 256$ & - \\
% \hline
% blocks5_2 ~ blocks5_6
5 $\times$ DepthSepConv  & $5\times 5$ & 1  & $14^2 \times 256$  & $14^2 \times 256$  & - \\
% blocks6_1
DepthSepConv      & $5\times 5$     & 2 & $14^2 \times 256$     & $7^2 \times 512$  & \checkmark \\ 
% \hline
% blocks6_2
DepthSepConv      & $5\times 5$     & 1 & $7^2 \times 512$      & $7^2 \times 512$  & \checkmark \\ 
% gap
GAP      & $7\times 7$     & 1 & $7^2 \times 512$      & $1^2 \times 512$  & - \\ 
%last-conv
Conv2d, NBN      & $1\times 1$     & 1 & $1^2 \times 512$      & $1^2 \times 1280$  & - \\ 
\bottomrule[1pt]
\end{tabular}
\end{center}

\caption{Architecture details of PP-LCNet. SE denotes
whether there is a Squeeze-and-Excitation in that block. NBN denotes no batch normalization.}
\label{tabimageclassification}
\end{table*}

While there are many lightweight networks whose inference speed is fast on ARM-based devices, few networks take into account the speed on Intel CPU, especially when acceleration strategies such as MKLDNN enabled. Many methods to improve model accuracy will not increase the inference time much on ARM devices, however, when 
switching to Intel CPU devices, the situation will be a little different. Here we have summarized some methods that can improve the performance of the model with little increase of inference time. These methods will be described in details below. We used the DepthSepConv mentioned by MobileNetV1\cite{mobilenetv1} as our basic block. This block does not have operations such as shortcuts, so there are no additional operations such as concat or elementwise-add, these operations will not only slow down the inference speed of the model, but also will not improve the accuracy on a small model. 
Furthermore, this block has been deeply optimized by the Intel CPU acceleration library, and the inference speed can surpass other lightweight blocks such as inverted-block or shufflenet-block. We stack these blocks to form a BaseNet similar to MobileNetV1\cite{mobilenetv1}. We combine the BaseNet and some of the existing technologies to a more powerful network, namely PP-LCNet.

\subsection{Better activation function}

As we all know, the quality of the activation function often determines the performance of the network. Since the activation function of network is changed from Sigmoid to ReLU, the performance of the network has been greatly improved. In recent years, more and more activation functions have emerged that go beyond ReLU. After EfficientNet\cite{efficientnet} used the Swish activation function to show better performance, the author of MobileNetV3\cite{mbv3} upgraded it to H-Swish, thus avoiding a large number of exponential operations. 
Since then, many lightweight networks also use this activation function. We also replaced the activation function in BaseNet from ReLU to H-Swish. The performance has been greatly improved, while the inference time has hardly changed.

\subsection{SE modules at appropriate positions}

The SE module\cite{senet} has been used by a large number of networks since its being proposed. This module also helped SENet\cite{senet} winning the 2017 ImageNet\cite{imagenet} classification competition. It does a good job of weighting the network channels for better features, and its speed improvement version is also used in many lightweight networks such as MobileNetV3\cite{mbv3}. However, on Intel CPUs, the SE module\cite{senet} increases the inference time, so that we cannot use it for the whole network. In fact, we have done a lot of experiments and observed that when the SE module\cite{senet} is located at the end of the network, it can play a better role. So we just add the SE module\cite{senet} to the blocks near the tail of the network. This results in a better accuracy-speed balance. As with MobileNetV3\cite{mbv3}, the activation functions for the two layers of the SE module\cite{senet} are ReLU and H-Sigmoid respectively.

\subsection{Larger convolution kernels}
The size of the convolution kernel often affects the final performance of the network. In MixNet\cite{mixnet}, the authors analysed the effect of differently sized convolution kernels on the performance of the network, and ended up mixing different sizes of convolutional kernels in the same layer of the network.  However, such a mixture slows down the inference speed of the model,  so we try to use only one size of convolution kernel in the single layer, and ensure that a large convolution kernel is used in the case of low latency and high accuracy. We experimentally find that, similar to the placement of the SE module\cite{senet}, replacing the $3\times3$ convolutional kernels with only the $5\times5$ convolutional kernels at the tail of the network would achieve the effect of replacing almost all layers of the network, so we did this replacement operation only at the tail of the network.

\begin{table*}[!htbp]
\centering
\begin{center}
\begin{tabular}{c|c|c|c|c|c}
\toprule[1pt]
Model & Params(M) & FLOPs(M) & Top-1 Acc.(\%) & Top-5 Acc.(\%) & Latency(ms) \\
\midrule[1pt]
PP-LCNet 0.25x     & 1.5     & 18     & 51.86	 & 75.65 & 1.74 \\
PP-LCNet-0.35x     &  1.6    & 29     & 58.09	 & 80.83 & 1.92 \\
PP-LCNet-0.5x     & 1.9     & 47    & 63.14	 & 84.66 & 2.05 \\
PP-LCNet-0.75x     & 2.4     & 99     & 68.18	 & 88.30 & 2.29 \\
PP-LCNet-1x     & 3.0     &161     & 71.32	 & 90.03 & 2.46 \\
PP-LCNet-1.5x     & 4.5     & 342     & 73.71	 & 91.53 & 3.19 \\
PP-LCNet-2x     & 6.5     & 590     & 75.18	 & 92.27 & 4.27 \\
PP-LCNet-2.5x     & 9.0     & 906     & 76.60	 & 93.00 & 5.39 \\
\hline
\textbf{PP-LCNet-0.5x*}     & \textbf{1.9}     & \textbf{47}    & \textbf{66.10}	 & \textbf{86.46} & \textbf{2.05} \\
\textbf{PP-LCNet-1x*}     & \textbf{3.0}     &\textbf{161}     & \textbf{74.39}	 & \textbf{92.09} & \textbf{2.46} \\
\textbf{PP-LCNet-2.5x*}     & \textbf{9.0}     & \textbf{906}     & \textbf{80.82}	 & \textbf{95.33} & \textbf{5.39} \\
\bottomrule[1pt]
\end{tabular}
\end{center}

\caption{Indicators of PP-LCNet of different scales, where * means it is trained using SSLD\cite{cui2021beyond} distillation method. Latency tested on Intel$^\circledR$ Xeon$^\circledR$ Gold 6148 Processor with batch size of 1 and  MKLDNN enabled, the number of thread is 10.}
\label{tab-acc-of-PP-LCNet}
\end{table*}

\begin{table*}[!htbp]
\centering
\begin{center}
\begin{tabular}{c|c|c|c|c|c}
\toprule[1pt]
Model & Params(M) & FLOPs(M) & Top-1 Acc.(\%) & Top-5 Acc.(\%) & Latency(ms) \\
\midrule[1pt]
MobileNetV2-0.25x     & 1.5     & 34     & 53.21	 & 76.52 & 2.47 \\
MobileNetV3-small-0.35x    & 1.7     & 15     & 53.03 & 76.37  & 3.02 \\
ShuffleNetV2-0.33x & 0.6     & 24     & 53.73  & 77.05  & 4.30 \\

\textbf{PP-LCNet-0.25x}    & \textbf{1.5}    & \textbf{18}    & \textbf{51.86}  & \textbf{75.65} & \textbf{1.74} \\
\hline
MobileNetV2-0.5x      & 2.0     & 99     & 65.03 & 85.72 & 2.85 \\
MobileNetV3-large-0.35x & 2.1 & 41  & 64.32  & 85.46  & 3.68 \\
ShuffleNetV2-0.5x     & 1.4     & 43 & 60.32    & 82.26  & 4.65 \\ 
\textbf{PP-LCNet-0.5x} & \textbf{1.9}  & \textbf{47}   & \textbf{63.14}      & \textbf{84.66}  &\textbf{2.05}\\ 

\hline
MobileNetV1-1x  & 4.3   & 578     & 	70.99 & 89.68 & 3.38 \\
MobileNetV2-1x  &  3.5  & 327     & 72.15 & 90.65 & 4.26 \\
MobileNetV3-small-1.25x & 3.6 & 100  & 70.67  & 89.51  & 3.95 \\
ShuffleNetV2-1.5x   & 3.5   & 301 & 71.63    & 90.15  & - \\ 
\textbf{PP-LCNet-1x}   & \textbf{3.0}  & \textbf{161}  & \textbf{71.32}    & \textbf{90.03}    &\textbf{2.46}\\ 

\bottomrule[1pt]
\end{tabular}
\end{center}

\caption{Comparison of state-of-the-art light networks over classification accuracy. Latency tested on Intel$^\circledR$ Xeon$^\circledR$ Gold 6148 Processor with batch size of 1 and  MKLDNN enabled, the number of thread is 10.}
\label{tab-state-of-the-art}
\end{table*}

\subsection{Larger dimensional $1 \times 1$ conv layer after GAP}

In our PP-LCNet, the output dimension of the network after GAP is small. And directly appending the final classification layer will lose the combination of features. In order to give the network a stronger fitting ability, we appended a 1280-dimensional size $1 \times 1$ conv(equivalent to FC layer) after the final GAP layer, which would allow for more storage of the model with little increase of inference time.

With these four changes, our model performs well on the ImageNet-1k\cite{imagenet}, and table \ref{tab-state-of-the-art} lists the metrics against other lightweight models on Intel CPUs.

\section{Experiment}

\subsection{Implementation Details}
For fair comparsions, we reimplement the models of MobileNetV1\cite{mobilenetv1}, MobileNetV2\cite{mobilenetv2}, MobileNetV3\cite{mbv3}, ShuffleNetV2\cite{shufflenetv2}, PicoDet\cite{ppdet2019} and Deeplabv3+\cite{deeplabv3+} by PaddlePaddle. We train the models on 4 V100 GPUs, and the CPU test environment is based on Intel$^\circledR$ Xeon$^\circledR$ Gold 6148 Processor with batch size of 1 and MKLDNN enabled.

\subsection{Image Classification}

For the image classification task, we train PP-LCNet on ImageNet-1k\cite{imagenet}, which contains 1.28 million training images and 50k validation images of 1000 classes. We use SGD optimizer with weight decay set to 3e-5 (4e-5 for large models), momentum set to 0.9, and batch size of 2048. Learning rate is adjusted according to a cosine schedule for training 360 epochs with 5 linear warmup epochs. Initial learning rate is set to 0.8.  In the training phase, each image is randomly cropped to $224\times224$ and randomly flipped horizontally. In the evaluation phase, we first resize an image to $256$ along the short edge, then apply a center crop of size $224 \times 224$. Table \ref{tab-acc-of-PP-LCNet} shows the PP-LCNet's top-1 and top-5 validation accuracy and inference time of different scales. Furthermore, when the SSLD\cite{cui2021beyond} distillation method is used, the accuracy of the model can be greatly improved. Table \ref{tab-state-of-the-art} shows the comparison of PP-LCNet and state-of-the-art models. Compared with other light models, PP-LCNet has shown strong competitiveness.

\subsection{Object Detection}

For object detection task, all models in Table \ref{tab:objectdetect} are trained on COCO-2017\cite{mscoco} training set with 80 classes and 118k images, and evaluated on COCO-2017\cite{mscoco} validation set with 5000 images using the common COCO AP metric of a single scale.  We used the lightweight PicoDet developed by PaddleDection\footnote{https://github.com/PaddlePaddle/PaddleDetection} as our baseline method. Table \ref{tab:objectdetect} shows the object detection results of PP-LCNet and MobileNetV3\cite{mbv3} as the backbone. The entire network is trained with stochastic gradient descent (SGD) for 146K iterations with a minibatch of 224 images distributed on 4 GPUs. The learning rate schedule is cosine from 0.3 as base learning rate for 280 epochs. Weight decay is set as 1e-4, and momentum is set as 0.9. Impressively, the PP-LCNet backbone greatly improves the mAP on COCO\cite{mscoco} and inference speed compared with MobileNetV3\cite{mbv3}.

\begin{table}[!htb]
%\centering
\begin{center}
\begin{tabular}{c|c|c|c}
% \hline
\toprule[1pt]
% \toprule
Method & Backbone & \makecell[c]{mAP\\(\%)} & \makecell[c]{Latency\\(ms)} \\
% \hline
\midrule[1pt]
\multirow{4}{*}{\shortstack{PicoDet}} & MobileNetV3-large-0.35x\cite{mbv3} & 19.2 & 8.1 \\
& \textbf{PP-LCNet-0.5x} & \textbf{20.3} & \textbf{6.0} \\
\cline{2-4}
& MobileNetV3-large-0.75x\cite{mbv3} & 25.8 & 11.1 \\
& \textbf{PP-LCNet-1x} & \textbf{26.9} & \textbf{7.9} \\
% \hline
\bottomrule[1pt]
\end{tabular}
\end{center}

\caption{Object detection results on the COCO dataset\cite{mscoco}, measured using mAP@IoU=0.5:0.95 (\%). Latency tested on Intel$^\circledR$ Xeon$^\circledR$ Gold 6148 Processor with batch size of 1 and  MKLDNN enabled, the number of thread is 10.}
\label{tab:objectdetect}
\end{table}

\subsection{Semantic Segmentation}

For the semantic segmentation task, we also evaluate the ability of PP-LCNet on Cityscapes dataset\cite{cityscapes}, which contains 5000 high-quality labeled images. We use the DeeplabV3+\cite{deeplabv3+} developed by PaddleSeg\footnote{https://github.com/PaddlePaddle/PaddleSeg} as our baseline method, and set the output stride to 32. The data are augmented by randomly horizontally flip, randomly scale, and randomly crop. The random scales contain $\{0.5, 0.75, 1.0, 1.25, 1.5, 1.75, 2.0\}$, and the cropped resolutions are $1024 \times 512$. We use the SGD optimizer with the initial learning rate of 0.01, the momentum of 0.9, and the weight decay of 4e-5. We use a poly learning rate schedule with a power of 0.9. All the models are trained for 80K iterations with the batch-size of 32 on 4 V100 GPUs.

We use MobileNetV3\cite{mbv3} as backbone for comparison. As shown in Table \ref{tab:semantic}, PP-LCNet-0.5x outperforms MobileNetV3-large-0.5x\cite{mbv3} by 2.94\% on mIoU, but the inference time is reduced by 53ms. Compared with larger models, PP-LCNet also has strong performance. When PP-LCNet-1x is used as backbone,  mIOU of model is 1.5\% higher than MobileNetV3-large-0.75x, but the inference time is reduced by 55ms.

\begin{table}[!htbp]
\centering
\begin{center}

\begin{tabular}{c|c|c|c}
\hline
\toprule[1pt]
 Method & Backbone & \makecell[c]{mIoU\\(\%)} & \makecell[c]{Latency\\(ms)}\\
\midrule[1pt]
\multirow{4}{*}{\shortstack{Deeplabv3+\cite{deeplabv3+}}}& \makecell[c]{MobileNetV3\\-large-0.5x\cite{mbv3}} & 55.42 & 135  \\
& \textbf{PP-LCNet-0.5x}   & \textbf{58.36}  & \textbf{82} \\
\cline{2-4}
& \makecell[c]{MobileNetV3\\-large-0.75x\cite{mbv3}}  & 64.53 & 151 \\
& \textbf{PP-LCNet-1x}   & \textbf{66.03}  & \textbf{96} \\

\bottomrule[1pt]
\end{tabular}
\end{center}

\caption{Performances of semantic segmentation on Cityscapes\cite{cityscapes} validation dataset. Latency tested on Intel$^\circledR$ Xeon$^\circledR$ Gold 6148 Processor with batch size of 1 and  MKLDNN enabled, the number of thread is 10.}
\label{tab:semantic}
\end{table}

\begin{table*}
%\centering
\begin{center}

\begin{tabular}{c|c|c|c|c|c}
% \hline
\toprule
 Activation & SE block  & large-kernel & last-1x1 conv & Top-1 Acc(\%)  & Latency(ms)\\
\midrule[1pt]
\XSolidBrush  & \checkmark  & \checkmark & \checkmark    & 61.93    & 1.94\\
\checkmark  & \XSolidBrush  & \checkmark & \checkmark    & 62.51    & 1.87\\
\checkmark  & \checkmark  & \XSolidBrush & \checkmark    & 62.44    & 2.01\\
\checkmark  & \checkmark  & \checkmark & \XSolidBrush   & 59.91    & 1.85\\
\hline 
\checkmark  & \checkmark  & \checkmark & \checkmark    & 63.14    & 2.05\\
\bottomrule
\hline
\end{tabular}
\end{center}

\caption{The impact of PP-LCNet-0.5x’s performance on reducing a certain technology. Latency tested on Intel$^\circledR$ Xeon$^\circledR$ Gold 6148 Processor with batch size of 1 and  MKLDNN enabled, the number of thread is 10.}
\label{tab:reduce-technologies}
\end{table*}

\subsection{Ablation Study}

\textbf{The impact of SE module\cite{senet} in different positions.} The SE module\cite{senet} is an attention mechanism between channels, which can improve the accuracy of the model. However, if the number of SE modules\cite{senet} is blindly increased, the inference speed of the model will be reduced, so it is worth studying and exploring how to properly add SE modules\cite{senet} to the model. Through experiments, we found that the SE module\cite{senet} will have a greater impact on the tail of the network. The results of adding only two SE modules\cite{senet} at different locations in the network are presented in the table \ref{tab:SE_ablation}. The table clearly shows that adding the last two blocks is more advantageous for almost the same inference time. Therefore, in order to balance the inference speed, PP-LCNet only adds the SE module\cite{senet} to the last two blocks.

\begin{table}[!htbp]
\centering
\begin{center}

\begin{tabular}{c|c|c|c}
% \hline
\toprule
 Network & SE Location & \makecell[c]{Top-1 Acc\\(\%)}& \makecell[c]{Latency\\(ms)}\\
\midrule[1pt]
\multirow{4}{*}{\shortstack{PP-LCNet-0.5x}}
& 1100000000000   & 61.73 &  2.06  \\
& 0000001100000   & 62.17 &  2.03  \\
& \textbf{0000000000011} & \textbf{63.14} &  \textbf{2.05}  \\
& 1111111111111   & 64.27 &  3.80  \\

\bottomrule[1pt]
\end{tabular}
\end{center}

\caption{Ablation experiment of SE module in different positions. Latency tested on Intel$^\circledR$ Xeon$^\circledR$ Gold 6148 Processor with batch size of 1 and  MKLDNN enabled, the number of thread is 10.}
\label{tab:SE_ablation}
\end{table}

\textbf{The impact of large-kernel in different locations}. Although large-kernel can increase accuracy, it is not the best to add it at all locations in the network. We have shown the general rule of correctly adding large-kernel through experiments. Table \ref{tab:large-kearnel_ablation} shows the positions added by the $5 \times 5$ depth-wise convolution. 1 means that the depth-wise convolution kernel in DepthSepConv is $5 \times 5$, and 0 means that the depth-wise convolution kernel in DepthSepConv is $3 \times 3$. It can be seen from the table that, similar to the location where the SE module\cite{senet} is added, the addition of $5 \times 5$ convolution at the tail of the network is also more competitive. Our PP-LCNet chose the configuration in the third row of the table.

\begin{table}[!htbp]
\centering
\begin{center}

\begin{tabular}{c|c|c|c}
% \hline
\toprule
 Network & \makecell[c]{Large-kernel\\location} & \makecell[c]{Top-1 Acc\\(\%)}& \makecell[c]{Latency\\(ms)}\\
\midrule[1pt]
\multirow{3}{*}{\shortstack{PP-LCNet-0.5x}}
& 1111111111111   & 63.22 &  2.08  \\
& 1111111000000   & 62.70 &  2.07  \\
& \textbf{0000001111111} & \textbf{63.14} &  \textbf{2.05}  \\

\bottomrule[1pt]
\end{tabular}
\end{center}

\caption{The impact of large-kernel in different locations.Latency tested on Intel$^\circledR$ Xeon$^\circledR$ Gold 6148 Processor with batch size of 1 and  MKLDNN enabled, the number of thread is 10.}
\label{tab:large-kearnel_ablation}
\end{table}

\textbf{The impact of  different techniques}.
In PP-LCNet, we use 4 different technologies to improve the performance of the model. Table \ref{tab:increase-technologies} lists the cumulative increase of different technologies on PP-LCNet, and Table \ref{tab:reduce-technologies} lists the impact of reducing different modules on PP-LCNet. It can be seen from the two tables that H-Swish and large-kernel can improve the performance of the model with almost no increase in inference time. Adding a small number of SE modules\cite{senet} can further improve the performance of the model. Using a larger FC layer after GAP will also greatly increase the accuracy. At the same time, perhaps because a relatively large matrix is involved here, the use of the dropout strategy can further improve the accuracy of the model.

\begin{table}[!htb]
%\centering
\begin{center}

\begin{tabular}{c|c|c}
% \hline
\toprule
 Strategy &  Top-1 Acc.(\%)  & Latency(ms)\\
\midrule[1pt]
BaseNet    & 55.58   & 1.61\\
+h-swish    & 58.18   & 1.66\\
+large-kernel    & 	59.09   & 1.70\\
+SE    & 59.91   & 1.85\\
+last-1x1 conv w/o dropout  & 62.50   & 2.05\\
\textbf{+last-1x1 conv w/ dropout}    & 	\textbf{63.14}   & \textbf{2.05}\\

\bottomrule
% \hline
\end{tabular}
\end{center}

\caption{The impact of the increase of different technologies on the performance of PP-LCNet-0.5x. Latency tested on Intel$^\circledR$ Xeon$^\circledR$ Gold 6148 Processor with batch size of 1 and  MKLDNN enabled, the number of thread is 10.}
\label{tab:increase-technologies}
\end{table}

\section{Conclusion and Future work}
Our work summarizes some methods for designing lightweight Intel CPU networks, which can improve the accuracy of the model while avoiding increasing the inference time. While these methods are existing methods from previous work, the balance between accuracy and speed has not been summarised experimentally. Through extensive experiments and blessing of these methods, we propose PP-LCNet, which shows stronger performance on a large number of vision tasks and has a better accuracy-speed balance. In addition, this work reduces the search space of NAS and also offers the possibility of faster access to lightweight models for NAS. In the future, we will also use NAS to obtain faster and stronger models.

%-------------------------------------------------------------------------

{\small
\bibliographystyle{unsrt}
\bibliography{egbib}

\begin{thebibliography}{10}

\bibitem{alexnet}
Alex Krizhevsky, Ilya Sutskever, and Geoffrey~E Hinton.
\newblock Imagenet classification with deep convolutional neural networks.
\newblock In {\em Advances in neural information processing systems}, pages
  1097--1105, 2012.

\bibitem{urnet}
Jia Li, Yafei Song, Jianfeng Zhu, Lele Cheng, Ying Su, Lin Ye, Pengcheng Yuan,
  and Shumin Han.
\newblock Learning from large-scale noisy web data with ubiquitous reweighting
  for image classification.
\newblock {\em IEEE Transactions on Pattern Analysis and Machine Intelligence},
  2019.

\bibitem{fasterrcnn}
Shaoqing Ren, Kaiming He, Ross Girshick, and Jian Sun.
\newblock Faster r-cnn: Towards real-time object detection with region proposal
  networks.
\newblock In {\em Advances in neural information processing systems}, pages
  91--99, 2015.

\bibitem{attentiongrad}
Ramprasaath~R Selvaraju, Michael Cogswell, Abhishek Das, Ramakrishna Vedantam,
  Devi Parikh, and Dhruv Batra.
\newblock Grad-cam: Visual explanations from deep networks via gradient-based
  localization.
\newblock In {\em Proceedings of the IEEE international conference on computer
  vision}, pages 618--626, 2017.

\bibitem{targettracking}
Tianzhu Zhang, Changsheng Xu, and Ming-Hsuan Yang.
\newblock Multi-task correlation particle filter for robust object tracking.
\newblock In {\em Proceedings of the IEEE conference on computer vision and
  pattern recognition}, pages 4335--4343, 2017.

\bibitem{action}
Karen Simonyan and Andrew Zisserman.
\newblock Two-stream convolutional networks for action recognition in videos.
\newblock In {\em Advances in neural information processing systems}, pages
  568--576, 2014.

\bibitem{deeplab}
Liang-Chieh Chen, George Papandreou, Iasonas Kokkinos, Kevin Murphy, and Alan~L
  Yuille.
\newblock Deeplab: Semantic image segmentation with deep convolutional nets,
  atrous convolution, and fully connected crfs.
\newblock {\em IEEE transactions on pattern analysis and machine intelligence},
  40(4):834--848, 2017.

\bibitem{deeplabv3+}
Liang-Chieh Chen, George Papandreou, Florian Schroff, and Hartwig Adam.
\newblock Rethinking atrous convolution for semantic image segmentation.
\newblock {\em arXiv preprint arXiv:1706.05587}, 2017.

\bibitem{SOD}
Ali Borji, Ming-Ming Cheng, Qibin Hou, Huaizu Jiang, and Jia Li.
\newblock Salient object detection: A survey.
\newblock {\em Computational visual media}, pages 1--34, 2019.

\bibitem{edgedetect}
Yun Liu, Ming-Ming Cheng, Xiaowei Hu, Kai Wang, and Xiang Bai.
\newblock Richer convolutional features for edge detection.
\newblock In {\em Proceedings of the IEEE conference on computer vision and
  pattern recognition}, pages 3000--3009, 2017.

\bibitem{zoph2016neural}
Barret Zoph and Quoc~V Le.
\newblock Neural architecture search with reinforcement learning.
\newblock {\em arXiv preprint arXiv:1611.01578}, 2016.

\bibitem{VGG}
Karen Simonyan and Andrew Zisserman.
\newblock Very deep convolutional networks for large-scale image recognition.
\newblock {\em arXiv preprint arXiv:1409.1556}, 2014.

\bibitem{inceptionv1}
Christian Szegedy, Wei Liu, Yangqing Jia, Pierre Sermanet, Scott Reed, Dragomir
  Anguelov, Dumitru Erhan, Vincent Vanhoucke, and Andrew Rabinovich.
\newblock Going deeper with convolutions.
\newblock In {\em Proceedings of the IEEE conference on computer vision and
  pattern recognition}, pages 1--9, 2015.

\bibitem{mobilenetv1}
Andrew~G Howard, Menglong Zhu, Bo~Chen, Dmitry Kalenichenko, Weijun Wang,
  Tobias Weyand, Marco Andreetto, and Hartwig Adam.
\newblock Mobilenets: Efficient convolutional neural networks for mobile vision
  applications.
\newblock {\em arXiv preprint arXiv:1704.04861}, 2017.

\bibitem{mobilenetv2}
Mark Sandler, Andrew Howard, Menglong Zhu, Andrey Zhmoginov, and Liang-Chieh
  Chen.
\newblock Mobilenetv2: Inverted residuals and linear bottlenecks.
\newblock In {\em Proceedings of the IEEE conference on computer vision and
  pattern recognition}, pages 4510--4520, 2018.

\bibitem{shufflenet}
Xiangyu Zhang, Xinyu Zhou, Mengxiao Lin, and Jian Sun.
\newblock Shufflenet: An extremely efficient convolutional neural network for
  mobile devices.
\newblock In {\em Proceedings of the IEEE conference on computer vision and
  pattern recognition}, pages 6848--6856, 2018.

\bibitem{shufflenetv2}
Ningning Ma, Xiangyu Zhang, Hai-Tao Zheng, and Jian Sun.
\newblock Shufflenet v2: Practical guidelines for efficient cnn architecture
  design.
\newblock In {\em Proceedings of the European conference on computer vision
  (ECCV)}, pages 116--131, 2018.

\bibitem{ghostnet}
Kai Han, Yunhe Wang, Qi~Tian, Jianyuan Guo, Chunjing Xu, and Chang Xu.
\newblock Ghostnet: More features from cheap operations.
\newblock In {\em Proceedings of the IEEE/CVF Conference on Computer Vision and
  Pattern Recognition}, pages 1580--1589, 2020.

\bibitem{efficientnet}
Mingxing Tan and Quoc~V Le.
\newblock Efficientnet: Rethinking model scaling for convolutional neural
  networks.
\newblock {\em arXiv preprint arXiv:1905.11946}, 2019.

\bibitem{mbv3}
Andrew Howard, Mark Sandler, Grace Chu, Liang-Chieh Chen, Bo~Chen, Mingxing
  Tan, Weijun Wang, Yukun Zhu, Ruoming Pang, Vijay Vasudevan, et~al.
\newblock Searching for mobilenetv3.
\newblock In {\em Proceedings of the IEEE International Conference on Computer
  Vision}, pages 1314--1324, 2019.

\bibitem{fbnet}
Bichen Wu, Xiaoliang Dai, Peizhao Zhang, Yanghan Wang, Fei Sun, Yiming Wu,
  Yuandong Tian, Peter Vajda, Yangqing Jia, and Kurt Keutzer.
\newblock Fbnet: Hardware-aware efficient convnet design via differentiable
  neural architecture search.
\newblock In {\em Proceedings of the IEEE Conference on Computer Vision and
  Pattern Recognition}, pages 10734--10742, 2019.

\bibitem{dnanet}
Changlin Li, Jiefeng Peng, Liuchun Yuan, Guangrun Wang, Xiaodan Liang, Liang
  Lin, and Xiaojun Chang.
\newblock Block-wisely supervised neural architecture search with knowledge
  distillation.
\newblock In {\em Proceedings of the IEEE/CVF Conference on Computer Vision and
  Pattern Recognition}, pages 1989--1998, 2020.

\bibitem{ofanet}
Han Cai, Chuang Gan, Tianzhe Wang, Zhekai Zhang, and Song Han.
\newblock Once-for-all: Train one network and specialize it for efficient
  deployment.
\newblock {\em arXiv preprint arXiv:1908.09791}, 2019.

\bibitem{mixnet}
Mingxing Tan and Quoc~V Le.
\newblock Mixconv: Mixed depthwise convolutional kernels.
\newblock {\em arXiv preprint arXiv:1907.09595}, 2019.

\bibitem{resnet}
Kaiming He, Xiangyu Zhang, Shaoqing Ren, and Jian Sun.
\newblock Deep residual learning for image recognition.
\newblock In {\em Proceedings of the IEEE conference on computer vision and
  pattern recognition}, pages 770--778, 2016.

\bibitem{senet}
Jie Hu, Li~Shen, and Gang Sun.
\newblock Squeeze-and-excitation networks.
\newblock In {\em Proceedings of the IEEE conference on computer vision and
  pattern recognition}, pages 7132--7141, 2018.

\bibitem{imagenet}
Jia Deng, Wei Dong, Richard Socher, Li-Jia Li, Kai Li, and Li~Fei-Fei.
\newblock Imagenet: A large-scale hierarchical image database.
\newblock In {\em 2009 IEEE conference on computer vision and pattern
  recognition}, pages 248--255. Ieee, 2009.

\bibitem{cui2021beyond}
Cheng Cui, Ruoyu Guo, Yuning Du, Dongliang He, Fu~Li, Zewu Wu, Qiwen Liu,
  Shilei Wen, Jizhou Huang, Xiaoguang Hu, et~al.
\newblock Beyond self-supervision: A simple yet effective network distillation
  alternative to improve backbones.
\newblock {\em arXiv preprint arXiv:2103.05959}, 2021.

\bibitem{ppdet2019}
PaddlePaddle Authors.
\newblock Paddledetection, object detection and instance segmentation toolkit
  based on paddlepaddle.
\newblock \url{https://github.com/PaddlePaddle/PaddleDetection}, 2019.

\bibitem{mscoco}
Tsung-Yi Lin, Michael Maire, Serge Belongie, James Hays, Pietro Perona, Deva
  Ramanan, Piotr Doll{\'a}r, and C~Lawrence Zitnick.
\newblock Microsoft coco: Common objects in context.
\newblock In {\em European conference on computer vision}, pages 740--755.
  Springer, 2014.

\bibitem{cityscapes}
Marius Cordts, Mohamed Omran, Sebastian Ramos, Timo Rehfeld, Markus Enzweiler,
  Rodrigo Benenson, Uwe Franke, Stefan Roth, and Bernt Schiele.
\newblock The cityscapes dataset for semantic urban scene understanding.
\newblock In {\em Proceedings of the IEEE conference on computer vision and
  pattern recognition}, pages 3213--3223, 2016.

\end{thebibliography}
}

\end{document}